\documentclass[letterpaper, 10 pt, conference]{ieeeconf}  

\IEEEoverridecommandlockouts                              

\usepackage{graphics} 
\usepackage{epsfig} 
\usepackage{times} 
\usepackage{amsmath} 
\usepackage{amssymb}  
\title{\LARGE \bf
Optimal feature rescaling in machine learning based on neural networks
}

\author{Federico Maria Vitrò$^{1}$ and Marco Leonesio$^{2}$ and Lorenzo Fagiano$^{3}$
\thanks{*This work was not supported by any organization.}
\thanks{$^{1}$Federico Maria Vitrò is a M.Sc. engineer in Automation and Control, Dipartimento di Elettronica, Informazione e Bioingegneria, Politecnico di Milano Piazza Leonardo da Vinci 32, 20133 Milano, Italy  
        {\tt\small federicomaria.vitro@mail.polimi.it}}%
\thanks{$^{2}$Marco Leonesio is researcher at the Institute of Intelligent Industrial Technologies and Systems for Advanced Manufacturing - National Research Council of Italy (STIIMA - CNR), via Corti 12, 20133 Milano, Italy 
        {\tt\small marco.leonesio@stiima.cnr.it}}%
\thanks{$^{3}$Lorenzo Fagiano is professor at Dipartimento di Elettronica, Informazione e Bioingegneria, Politecnico di Milano Piazza Leonardo da Vinci 32, 20133 Milano, Italy
        {\tt\small lorenzo.fagiano@polimi.it}}%
}

\begin{document}

\maketitle
\thispagestyle{empty}
\pagestyle{empty}

\begin{abstract}
This paper proposes a novel approach to improve the training efficiency and the generalization performance of Feed Forward Neural Networks (FFNNs) resorting to an optimal rescaling of input features (OFR) carried out by a Genetic Algorithm (GA). The OFR reshapes the input space improving the conditioning of the gradient-based algorithm used for the training. Moreover, the scale factors exploration entailed by GA trials and selection corresponds to different initialization of the first layer weights at each training attempt, thus realizing a multi-start global search algorithm (even though restrained to few weights only) which fosters the achievement of a global minimum. The approach has been tested on a FFNN modeling the outcome of a real industrial process (centerless grinding).

\end{abstract}


\section{INTRODUCTION}
Training in Neural Networks (NNs) is a very delicate task that plays a paramount role in determining the final model performance \cite{Shang96}, while raising efficiency issues in terms of computational time \cite{Wan2019}. Supervised training can be considered an unconstrained nonlinear minimization problem where the objective function is defined by a loss function depending on NN weights. Unfortunately, the loss function is usually non-convex and irregular, exhibiting several local minima, thus making the minimization problem rather hard. For these reasons, several approaches have been proposed in the literature to increase the efficiency of NNs training and to seek the global optimal weights set \cite{Prateek2017}.\par
As an alternative to the more traditional local search algorithms (gradient-based), some works have investigated the so-called global search algorithms, which can be divided into deterministic and stochastic \cite{Loc2021}. Deterministic global algorithms are characterized by significant computational requirements for problems with more than a few parameters and are usually avoided. Instead, stochastic global algorithms are more viable and efficient. Some examples of stochastic search algorithms are simulated annealing and various evolutionary algorithms such as genetic algorithms, evolutionary strategies, and evolutionary programming. However, despite the potential theoretical advantages of global search algorithms, local gradient-based algorithms are still the most widespread: in fact, restarts of a local search algorithm with random weights are a valid alternative to using global algorithms in finding global minima (actually, it can be assimilated to a stochastic global search algorithm) \cite{Hamm2002}.\par
On the other side, another important caveat of NNs is the training convergence time. Existing theoretical results are mostly negative, showing that successfully learning with these networks is computationally hard in the worst case \cite{Blum92}. In this perspective, the starting values of the weights can have a significant effect on the efficiency training process using gradient-based algorithms. For instance, in dealing with sigmoid-NN, weights can be chosen randomly but considering that the sigmoid should be primarily activated in its linear region. If weights are all very large, then the sigmoid saturates resulting in small gradients that make learning slow; complementary drawbacks occur if weights are too small \cite{Greg2012}. In general, faster convergence can be attained by normalizing the variance of input variables and/or associating them with different learning rates. Both objectives can be pursued by a proper input \textit{rescaling}.\par
On these premises, this paper proposes a novel approach to improve efficiently the generalization performance of Feed Forward Neural Networks (FFNNs) resorting to an optimal rescaling of input features (OFR) carried out by a Genetic Algorithm (GA). The OFR reshapes the input space improving the conditioning of the gradient-based algorithm used for the training (i.e., ADAM \cite{Kingma2014}). Moreover, the scale factors exploration entailed by GA trials and selection corresponds to different initialization of the first layer weights at each training attempt, thus realizing a multi-start global search algorithm (even though restrained to few weights only) which fosters the achievement of a global minimum. Finally, in the proposed approach the GA objective function is represented by the fitting error on the validation set, unlike the other usual approaches that consider the sole training error. \par 
The OFR efficacy has been tested on an FFNN modeling the outcome of a real industrial process (centerless grinding).

\section{PROPOSED METHOD:\\ OPTIMAL FEATURE RESCALING}

Feature rescaling is one of the most critical parts of the pre-processing phase in machine learning \cite{Ioffe2015, Greg2012}. In fact, it can make the difference between weak and strong machine learning models. In feature rescaling the problem variables are differently transformed depending on the method applied.
Usually, these techniques are used to:
\begin{itemize}
	\item make the features comparable in order to get a clearer insight about their relative importance/variability;
	\item make a few algorithms converge faster, e.g. training processes of neural networks.
\end{itemize}
In this work, features rescaling is proposed to improve FFNN performances not only in terms of training convergence velocity but also in generalization error.\par
The idea is to find a set of rescaling parameters $\mathbf{s}=\{s_1, ..., s_M\}$ that can be used to rescale the $M$ input variables of a regression problem. Namely, given a dataset $\mathbf{X}\in\mathbb{R}^{N \times M}$ and a vector of rescaling parameters $\mathbf{s}\in\mathbb{R}^{M}$, the rescaled dataset $\tilde{\mathbf{X}}\in\mathbb{R}^{N \times M}$ is defined as:

\begin{equation}
\label{eq:scaldef}
\tilde{\mathbf{X}}:=\mathbf{X}\mathbf{S}=
\begin{bmatrix}
			s_1 \cdot \begin{pmatrix} x_{11}  \\ \vdots \\ x_{N1} \end{pmatrix},
			&
			\cdots,
			&
			s_M \cdot \begin{pmatrix} x_{1M}  \\ \vdots \\ x_{NM} \end{pmatrix}
		\end{bmatrix}
\end{equation}

with $\mathbf{S}:=\text{diag}(s_i)$, and $\tilde{\boldsymbol{x}}_{i} := [s_ix_{1i}, ... , s_ix_{Ni}]^T,\; i=1, ..., M$ is the \textit{i}-th rescaled feature of the problem.

In order to improve the prediction capabilities of the model (in this study, a FFNN), a vector of optimal parameters $\mathbf{s}^*$ has to be chosen, hence the name "Optimal Feature Rescaling". Therefore, an optimization problem is to be posed.


\subsection{Optimization problem}
In order to set a global optimization problem an objective function has to be defined. In our case, the considered objective function takes as inputs the vector of features scaling factors (namely, the optimization problem \textit{decision variables}) and the training/validation sets. After having applied the features rescaling, the target FFNN is trained by fitting the training set and used to calculate the Root Mean Squared Error (RMSE) on the validation set. As the FFNN training implies a further optimization problem on the weights, the whole procedure corresponds to 2 nested optimization problems.\par
Let $\mathcal{D}_{trn}:=\{\mathbf{X}_{trn},\mathbf{y}_{trn}\}$ and $\mathcal{D}_{val}:=\{\mathbf{X}_{val},\mathbf{y}_{val}\}$ be the training and validation sets respectively. Then, the primal objective function to be minimized w.r.t. the scaling factor $\mathbf{s}$ can be defined as follows:
\begin{multline}
\label{eq:obj1}
	\mathrm{RMSE}_{val} := f(\mathbf{s}| \mathbf{W}(\mathcal{D}_{trn}), \mathcal{D}_{val})=
 \\
 =f(h_{\text{NN}}(\tilde{\mathbf{X}}_{val}|\mathbf{W}(\mathcal{D}_{trn})),\mathbf{y}_{val})=\\
 =\sqrt{\frac{1}{|\mathcal{D}_{val}|}\sum_{\mathcal{D}_{val}}\left(h_{\text{NN}}(\tilde{\mathbf{x}}_i|\mathbf{W}(\mathcal{D}_{trn}))^2-y_i^2 \right)}
\end{multline}
where $h_{\text{NN}}$ is a given FFNN parameterized w.r.t. the weights tensor $\mathbf{W}:=\{w^l_{jk}\}$. Note that in the last term of the (\ref{eq:obj1}) the dependence on $\mathbf{s}$ is implicitly considered by the rescaled features vector $\tilde{\mathbf{x}}_i$, as defined in the (\ref{eq:scaldef}).  On its turn, the optimal $\mathbf{W}$ can be obtained by solving the training optimization problem with the rescaled inputs exploiting the Mean Absolute Error (MAE) objective function, namely:
\begin{gather}
\label{eq:obj2}
	\mathrm{MAE}_{trn}:= f(\mathbf{W}|\mathbf{s}, \mathcal{D}_{trn})=\\
	=\frac{1}{|\mathcal{D}_{trn}|}\sum_{\mathcal{D}_{trn}}\left| h_{\text{NN}}(\tilde{\mathbf{x}}_i|\mathbf{W}(\mathcal{D}_{trn})-y_i \right|
\end{gather}

Therefore, the overall optimization program can be stated as follows:

\begin{equation}
    \begin{aligned}
    \label{ob:fin}
    &\min_\mathbf{s} \mathrm{RMSE}_{val}(\mathbf{s}| \mathbf{W}(\mathcal{D}_{trn}), \mathcal{D}_{val})\\
    \text{s.t.}\:\:\: &\mathbf{W}=\arg\min_\mathbf{W} \mathrm{MAE}_{trn}(\mathbf{W}|\mathbf{s}, \mathcal{D}_{trn})
\end{aligned}
\end{equation}
\par


An interesting observation highlights that the input rescaling equates to rescaling the weights of the first NN layer. Let $a^l_j$ be the $j$-th input to the layer $l$, and $w^l_{jk}$ the weight from the $k$-th neuron in the $(l-1)$-th layer to the $j$-th neuron in the $l$-th layer; the neuron output becomes:

\begin{equation}
    \label{eq:NN_weight}
	a^l_j=f(\sum_{k} w^l_{jk}a^{l-1}_k+b^l_j)
\end{equation}

where $b^l_j$ is the $j$-th bias to the layer $l$ and $f(\cdot)$ is the activation function of the considered layer. Then, applying rescaling to the first layer yields:
\begin{equation}
\label{eq:NN_weight2}
		a^1_j=f(\sum_{k} w^1_{jk}s_ka^{0}_k+b^1_j)=f(\sum_{k} \tilde{w}^1_{jk}a^{0}_k+b^1_j)
\end{equation}
with 
\begin{equation}
\label{eq:NN_weight3}
	\tilde{w}^1_{jk}:=w^1_{jk}s_k
\end{equation}

Going back to the optimization problem, it can be observed that the analytical form of the original training objective function is not known beforehand. Anyway, it can be stated that the objective function (\ref{ob:fin}) is non-convex. This claim comes from the empirical finding that, using backpropagation gradient-based methods, different scale factors make training converge to different minima. Whereas different scale factors correspond to different weights in the original input space (as shown by (\ref{eq:NN_weight2}) and (\ref{eq:NN_weight3})) and, then, different starting weights in the minimization algorithm, the resulting multiple local minima can be traced back to the non-convexity of the original problem. Indeed, if the original problem were convex, then its convexity would remain unchanged for any input rescaling - which is an affine transformation - due to the following theorem \cite{Boyd2004}:\\

\textit{Affine input transformation}. If $f:\Omega \rightarrow \mathbb{R}$ is convex (where $\Omega \subseteq \mathbb{R}^n$), then also $\tilde{f}(\mathbf{x})=f(A\mathbf{x}+\mathbf{b})$ is convex on the domain $\tilde{\Omega}=\{\mathbf{x} \in \mathbb{R}^m:A\mathbf{x}+b\in \Omega\}$, with $A\in\mathbb{R}^{n \times m}, \mathbf{b}\in\mathbb{R}^m$.\\\par

Once the optimization problem is set, a solver should be used in order to get to a solution. While for the high-dimensional inner training problem (unconstrained minimization of (\ref{eq:obj2})) an efficient local gradient-based method has been selected (ADAM), the primal optimization problem aimed at identifying the scale factors resorts to a \textit{Genetic Algorithm}, i.e., a stochastic global search algorithm, with the intent to tackle its inherent non-convexity. 


\subsection{Genetic Algorithm}
\label{sec:GA}
GA are metaheuristics that belong to the family of evolutionary algorithms; they are inspired by the natural evolutionary theory and try to emulate the process of natural selection where the next generation of offspring is produced by a group of individuals selected for reproduction \cite{EssMetaheuristic}. 
The GA implemented during this work follows 5 steps in order to provide a solution:
\begin{enumerate}
    \item \textit{Initialization}. The GA was initialized with a population of candidate solutions (the so called individuals), sampled from a uniform distribution between $[-3,3]$ in logarithmic scale. They were then converted in decimal scale, i.e. $ w \in [-10^3, 10^3]$, before being applied to the dataset features. The population size is set to 20, as suggested in \cite{DorseyMayer}. Each individual was then evaluated using the objective function keeping track of the best performance.
    \item \textit{Selection and crossover}. For a number of times equal to the desired number of offsprings (20 in the considered case) 2 individuals ($\mathbf{x}_1$ and $\mathbf{x}_2$) were randomly selected among the initial population in order to perform a crossover. Crossover is an operator used to combine the genetic information of 2 parents in order to generate 2 new offsprings ($\mathbf{y}_1$ and $\mathbf{y}_2$). In this case a \textit{uniform crossover} was implemented so, if $\mathbf{x}_1 = (x_{11}, x_{12},..., x_{1n}) \in \mathbb{R}^n$ and $\mathbf{x}_2 = (x_{21}, x_{22},..., x_{2n}) \in \mathbb{R}^n$ are the 2 selected parents and $\mathbf{y}_1 = (y_{11}, y_{12},..., y_{1n}) \in \mathbb{R}^n$, $\mathbf{y}_2 = (y_{21}, y_{22},..., y_{2n}) \in \mathbb{R}^n$ the offspring, the $i$-th element of each offspring was computed applying \eqref{eq:UniformCrossover1} - \eqref{eq:UniformCrossover2}:
    \begin{equation}
        \label{eq:UniformCrossover1}
        y_{1i} = \alpha x_{1i} + (1-\alpha)x_{2i}
    \end{equation}
        \begin{equation}
        \label{eq:UniformCrossover2}
        y_{2i} = \alpha x_{2i} + (1-\alpha)x_{1i}
    \end{equation}
    where $i=1,...,n$ and $\alpha$ is a parameter sampled from a uniform distribution between $[-0.1,0.1]$.
    
    \item \textit{Mutation}: the offspring calculated in the previous step were mutated with a mutation rate $\mu=0.2$ (i.e. each element of the considered offspring is mutated with a probability less or equal to $\mu$). The elements of the offspring are mutated following \eqref{eq:mutation}:
    \begin{equation}
        \label{eq:mutation}
         y^{'} = y + \mathcal{N}(0,\sigma^2)   
    \end{equation}
    where $y^{'}$ is the mutated element, $y$ is the original one and $\sigma = 0.1$. If a mutated element $y^{'} \notin [-3,3]$, it was replaced with the nearest extreme.

    \item \textit{Merge}. The new offspring were merged to the original population.

    \item \textit{Evaluate, sort \& select}. The new population was evaluated with the objective function and sorted. The best 20 individuals were maintained and the best performance was updated.
\end{enumerate}
These steps were repeated until the maximum number of iterations was reached. In this case, the GA was executed for 100 iterations, i.e. 2020 function evaluations.

\section{CASE STUDY}

\begin{figure}
    \centering
    \includegraphics[width=0.45\textwidth]{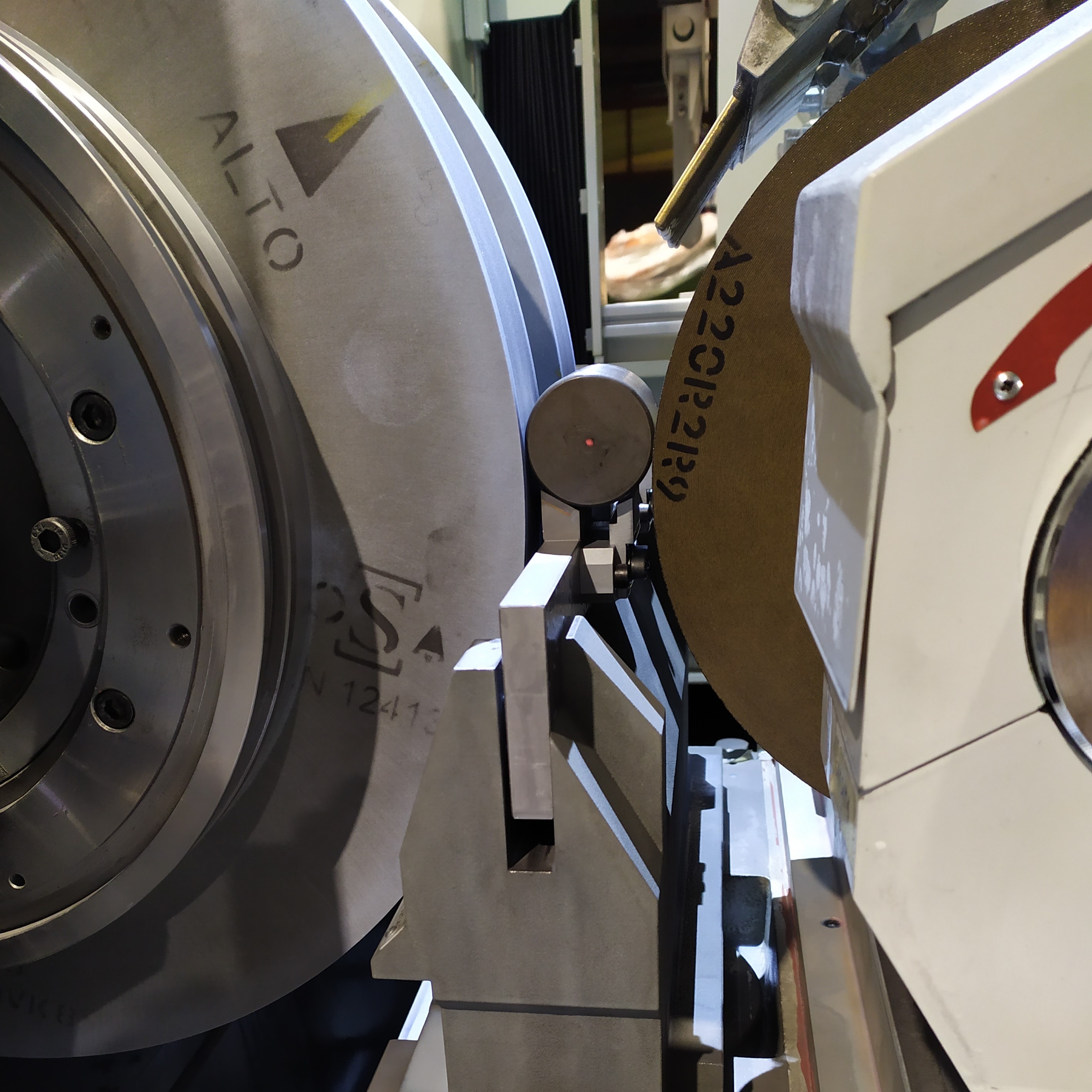}
    \caption{Centerless grinding process}
    \label{fig:CentGrind}
\end{figure}

The OFR method was tested in a real scenario: \textit{roundness prediction for a centerless grinding machining process}. Centerless grinding is a manufacturing process where a cylindrical component is ground without using a spindle or center to support the workpiece. Instead, it uses a regulating wheel and a support blade that holds and guides the workpiece during the grinding process (see Fig. \ref{fig:CentGrind}). This allows for a faster load/unload procedure in comparison with the traditional on-center grinding. Unfortunately, the fact that work part is not clamped during the machining process makes it prone to an inherent dynamic instability and, therefore, the overall quality of the final piece (essentially represented by its \textit{roundness}) is difficult to predict on the basis of the chosen process parameters set. \cite{Zhou96}.\par 

In order to support the operator in the centerless grinding setup, a decision supporting system relying on a model of the process is needed \cite{LeoIRIM}. A pure data-driven model would need too much data to correctly capture the non-smooth process behavior; on the other side, a complex physic-based may be non-robust, due to unavoidable model mismatch and parameters uncertainties. Hybrid approaches combining physics-based and data-driven techniques seems to perform better \cite{LeoSPIC}. On this track, the FFNN for roundness prediction considered in this study realizes such a hybridization by enhancing the features that define the grinding process with a further feature represented by the roundness prediction coming from a physic-based model.


\subsection{Dataset generation}
In the proposed centerless grinding model, each operation can be defined by 12 parameters which are listed in Table \ref{table:regression_input}.
\begin{table}
    \caption{Centerless grinding process parameters \cite{Rowe2014}.}
    \label{table:regression_input}
    \begin{center}
	   \begin{tabular}{ll}
	       \hline
    		$x_0$: Work height				& $x_6$: grinding wheel diameter \\
    		$x_1$: blade angle				& $x_7$: control wheel diameter \\
    		$x_2$: feed velocity 			& $x_8$: control wheel velocity \\
    		$x_3$: Diametral removal 		& $x_9$: grinding specific energy \\
    		$x_4$: Work length 				& $x_{10}$: edge force component \\
    		$x_5$: Work diameter 			& $x_{11}$: grit stiffness \\
    	   \hline
	   \end{tabular}
    \end{center}
\end{table}
The definition of the above-mentioned parameters can be found in all textbooks dealing with grinding technology, for instance \cite{Rowe2014}. The model requires the definition of other 9 parameters that are characteristic of the grinding machine and not of the process. They will be kept fixed and not considered in this study.\par
In absence of a proper experimetal dataset, a detailed model of the grinding process, the so-called \textit{high fidelity model}, is then used in order to obtain a virtual dataset for the regression problem: $\mathcal{D}:=\{y_i, \textbf{x}_i\}, i=1,2, ..., N_s$, where $y_i \in \mathbb{R}^+$ is the continuous value of the roundness of the final piece that measures the process performances of the $i$-th sampled grinding operation (i.e. the target variable), $\textbf{x}_i \in \mathbb{R}^{12}$ is the corresponding vector of process parameters (i.e. the input variables), and $N_s = 4069$ is the number of samples. A comprehensive description of the high-fidelity model can be retrieved in \cite{c10}.
In order to realize the physic-based hybridization, the feature set (that consisted of the 12 aforementioned parameters) is augmented by using the output of the so-called \textit{low fidelity model}, which is a simplified version of the high fidelity one, less accurate, but more handy and general (basically, it takes into account only kinematic aspects like rounding mechanism wheel-work detachment, neglecting work-machine relative dynamic compliance) \cite{c8}. 

The dataset thus obtained was divided in 3 subsets by applying a \textit{hold-out procedure}: training (70\%), validation (15\%) and test (15\%) sets.

In order to reserve the majority of the data for the FFNN training, only a small portion of them is held out for the test phase. For this reason, a \textit{10-folds cross-validation procedure} is used in order to test the performances of the FFNN using as a metric the \textit{Coefficient of determination} ($\mathrm{R^2}$). Because of the dimensions of the test set, it has been only used in cases where the cross-validation procedure is not a viable option.


\subsection{Test 1: OFR application}
\label{sec:test1}

Once the dataset is defined, the OFR approach is tested by training a FFNN on an optimally rescaled set of data.
The tested FFNN, i.e. the considered hypothesis space $H$, is composed by:
\begin{itemize}
	\item an input layer with 128 neurons and \textit{relu} activation function;
	\item 3 hidden layers with 256 neurons each and \textit{relu} activation functions;
	\item an output layer with a single neuron with \textit{linear} activation function.
\end{itemize}

MAE is used as loss metric both for the training and validation phases, while the optimization is based on ADAM algorithm. In the end, 500 \textit{epochs} have been considered during the training phase and the neural network is trained with all the training data in each one of them.

The FFNN described above is implemented in Python making use of Keras library.
Keras is a deep learning framework for Python that provides a practical way to define and train many different deep learning models \cite{DeepWithPython}.\par

Then, the considered FFNN iis trained 100 times on standardized data and the obtained performance are averaged and used to define a baseline for the test (BASE).

The baseline is compared with the mean performances obtained by training 100 FFNNs on the optimally scaled data, i.e. by applying OFR on previously standardized data (OFR).

The obtained results are reported in Table \ref{tab:perf_comparison_noES}.

\begin{table}
	\caption{Test 1: performances comparison ordered w.r.t. "CV Mean".}
	\label{tab:perf_comparison_noES}
    \begin{center}
    	\begin{tabular}{|c|c|c|c|}
    	\hline
    	   \textbf{Method} & \textbf{$\mathrm{R^2}$ (Train)} & \textbf{CV Mean} & \textbf{CV Std} \\
    	\hline \hline
    	   OFR & 0.808739 & 0.559659 & 0.133694 \\
    	\hline
    	   BASE & 0.990043 & 0.369556 & 0.337064 \\
    	\hline
    	\end{tabular}
    \end{center}
\end{table}

The performance improvement obtained by applying OFR to a standardized dataset is far greater than all of the other cases, guaranteeing an increment of 51.44\% compared to using the standardized data.\par 

Nevertheless, analyzing the discrepancy between the performances on the train set and the ones obtained with cross-validation, it can be deduced that the model is affected by the overfitting problem, so the \textit{Early Stopping} (ES) technique was applied in test 2. ES is a sort of regularization method used to avoid overfitting: it does not change the loss function, but just stops the training when the model is getting ready to learn noise. Such method works with iterative learning algorithms, such as gradient descent. It consists in monitoring the generalization error computed on the validation set and stopping the training error minimization when this error does not improve for a given iteration number (called \textit{patience}). 


\subsection{Test 2: OFR application with ES}
\label{sec:test2}
In order to counteract the overfitting problem, an ES approach is followed. The number of epochs for the FFNN training has been increased to 10000, while the ES patience is set to 100. The goal of this choice is to make sure that the ending of the training phase is reached before the end of the epochs. In this way, it is guaranteed that the FFNN training stops when the validation error begins to rise.

In this case the results, reported in Table \ref{tab:perf_comparison_siES}, are averaged with respect to the performances obtained by training and testing 10 FFNNs.

\begin{table}
\caption{Test 2: performances comparison ordered w.r.t. "CV Mean" (with ES).}
\label{tab:perf_comparison_siES}
    \begin{center}
    	\begin{tabular}{| c | c | c | c |}
    	\hline
    		\textbf{Method} & \textbf{$\mathrm{R^2}$ (Train)} & \textbf{CV Mean} & \textbf{CV Std} \\
    	\hline \hline
    		OFR & 0.806405 & 0.507067 & 0.145582 \\
    	\hline
    		BASE & 0.844916 & 0.432999 & 0.168303 \\
    	\hline
    	\end{tabular}
    \end{center}
\end{table}

The improvement achieved by using the OFR technique can be seen from Table \ref{tab:perf_comparison_siES}. However, the performances of the FFNN are still rather poor despite the implementation of ES. This probably depends on several factors:
\begin{enumerate}
	\item the FFNN is too complex for the considered problem;
	\item the parameters that were set for the ES were too permissive;
	\item the numerosity of the data;
	\item the high variance on the data.
\end{enumerate}
The prediction performances of the network would surely benefit from a \textit{model selection} procedure, where several models are tested on a validation set in order to choose the best among them.


\subsection{Test 3: OFR effects on a simplified FFNN with ES}
\label{sec:test3}
In order to test the efficacy of the OFR method with a model not subjected to the overfitting problem a simpler neural network has been tested. The FFNN described in Section \ref{sec:test1} (i.e. the hypothesis space $H$) is simplified by drastically reducing the number of layers and neurons. The resulting model is composed by:
\begin{itemize}
	\item an input layer with 13 neurons and \textit{relu} activation function;
	\item 1 hidden layer with 100 neurons and \textit{relu} activation function;
	\item an output layer with a single neuron with \textit{linear} activation function.
\end{itemize}
Furthermore, the objective function for the optimization program is changed, reflecting the new considered model. The ES patience is incremented to 200, in order to gain more training time for the FFNN.

As in the previous case, the results reported in Table \ref{tab:perf_comparison_simple_siES} are averaged with respect to the performances obtained by training and testing 10 FFNNs.

\begin{table}
\caption{Test 3: performances comparison obtained by training a simplified FFNN (with ES), ordered with respect to "CV Mean".}
\label{tab:perf_comparison_simple_siES}
    \begin{center}
    	\begin{tabular}{| c | c | c | c | c |}
    	\hline
    		\textbf{Method} & \textbf{$\mathrm{R^2}$ (Train)} & \textbf{CV Mean} & \textbf{CV Std} \\
    	\hline \hline
    		OFR & 0.627320 & 0.586836 & 0.121164 \\
    	\hline
    		BASE & 0.622829 & 0.562670 & 0.136008 \\
    	\hline
    	\end{tabular}
\end{center}
\end{table}

As it can be seen from Table \ref{tab:perf_comparison_simple_siES} the network trained on the optimally scaled dataset achieves better performances compared to the baseline. Furthermore, the overfitting problem is almost completely solved, significantly reducing the gap between the training error and the crossvalidated performances.


\subsection{Genetic Algorithm computational time}

The execution times of the genetic algorithm described in Section \ref{sec:GA}, for each of the 3 tests performed in Section \ref{sec:test1}, \ref{sec:test2} and \ref{sec:test3}, are reported in Table \ref{tab:ex_time}.

\begin{table}
\caption{GA execution times for test 1, 2 and 3.}
\label{tab:ex_time}
    \begin{center}
    	\begin{tabular}{|c|c|c|}
    	\hline
    		\textbf{Test} & \textbf{Time [s]} & \textbf{Time [h]} \\
    	\hline \hline
    		Test 3 & 134213.0 & 37.28 \\
    	\hline
    		Test 1 & 119886.0 & 33.30 \\
    	\hline
         	Test 2 & 61768.0 & 17.16 \\
    	\hline
    	\end{tabular}
\end{center}
\end{table}

As it can be seen from the Table \ref{tab:ex_time}, in test 3 the GA requires more time than the other 2 tests. This is certainly due to the high number of epochs (10000), but also to the patience chosen for the ES. In fact, in test 3, the latter is set to 200; therefore the network, although simpler than the others, requires more training time. As expected, test 1 requires the least computational time, since the network employed is trained using a smaller number of epochs (500) and does not foresee the use of the ES.


\subsection{Efficiency analysis}

As demonstrated in the previous paragraphs, the application of the OFR technique allows to obtain better performances rather than only applying standardization. In order to evaluate whether OFR also improves the efficiency of the NN training phase, 3 more tests are performed. For each of the previously tested NNs, the training time is calculated considering both the standardized dataset and the one to which, after the standardization phase, the OFR technique has been applied.
The results obtained are reported in the Tables \ref{tab:EffComp_test1}, \ref{tab:EffComp_test2}, \ref{tab:EffComp_test3}.

\begin{table}
\caption{Test 4.1: efficiency comparison using the network described in test 1.}
\label{tab:EffComp_test1}
    \begin{center}
        \begin{tabular}{|c|c|c|c|}
        	\hline
        		\textbf{Method} & \textbf{$\mathrm{R^2}$ (Train)} & \textbf{$\mathrm{R^2}$ (Test)} & \textbf{Time [s]} \\
        	\hline \hline
        		OFR & 0.842856 & 0.515479 & 74.447800\\
        	\hline
        		BASE &0.979330 & 0.386531 & 72.647971\\
        	\hline
        	\end{tabular}
        \end{center}
\end{table}

\begin{table}
\caption{Test 4.1: efficiency comparison using the network described in test 2.}
\label{tab:EffComp_test2}
    \begin{center}
            \begin{tabular}{| c | c | c |c|}
        	\hline
        		\textbf{Method} & \textbf{$\mathrm{R^2}$ (Train)} & \textbf{$\mathrm{R^2}$ (Test)} & \textbf{Time [s]}\\
        	\hline \hline
        		OFR & 0.799192 & 0.460403 & 25.518446\\
        	\hline
        		BASE & 0.820355 & 0.408107 & 21.681283\\
        	\hline
        	\end{tabular}
           \end{center}
\end{table}

\begin{table}
\caption{Test 4.1: efficiency comparison using the network described in test 3.}
\label{tab:EffComp_test3}
    \begin{center}
        \begin{tabular}{| c | c | c | c |}
        	\hline
        		\textbf{Method} & \textbf{$\mathrm{R^2}$ (Train)} & \textbf{$\mathrm{R^2}$ (Test)} & \textbf{Time [s]} \\
        	\hline \hline
        		OFR & 0.630621 & 0.497609 & 104.480643\\
        	\hline
        		BASE & 0.617340 & 0.366670 & 75.148524\\
        	\hline
        	\end{tabular}
       \end{center}
\end{table}

As it can be seen in Tables \ref{tab:EffComp_test1}, \ref{tab:EffComp_test2}, \ref{tab:EffComp_test3}, alongside an improvement in the generalization performances of NNs, the training time greatly increases by applying OFR together with classical standardization. This is probably due to the fact that, by only applying standardization, a shallower local minimum is obtained compared to the one reached by applying the OFR technique. So, although it takes less time to complete the training, the network is less performing.

\section{CONCLUSIONS}

In this paper, a new method to improve the performance of a Feed Forward Neural Network in regression problems has been defined and tested. This method is called \textit{Optimal Feature Rescaling} (OFR) and consists in multiplying the model variables, i.e. the features of the input dataset, by an optimal parameters set obtained by solving a global optimization problem. Unlike most of the current literature, our optimization problem focuses on the generalization error (i.e., the loss in validation set) and not the training error. Since the neural network training problem is usually non-convex, it has been shown that the optimal rescaling factors contributes in skip local minima, demonstrating the effectiveness of this approach in improving the FFNN performance. Moreover, tests have shown that it is possible to apply the OFR within the network, incorporating the optimal parameters in the first level weights. \par
The Optimal Features Rescaling approach was then applied to a real industrial scenario to test its efficacy: the roundness prediction for a centerless grinding machining process. The results demonstrated that OFR over-performs the current best practice, i.e., feature standardization.\par 
Despite the performance improvement, tests have shown that reaching a better minimum point also entails more training time. Therefore, the application of the OFR technique guarantees an improvement in performance at the cost of a lower training efficiency.\par
Finally, under the reasonable assumption that the OFR efficacy can be ascribed to the application of a sort of mutli-start optimization approach restrained to the first weights layer in FFNN training, future developments will be dedicated to evaluate the application of a similar global optimization approach for all the weights of the network.

\addtolength{\textheight}{-12cm}   


\section*{ACKNOWLEDGMENT}
The authors acknowledge Regione Lombardia and Monzesi spa who provided respectively the funding and the industrial application for the development of the research.  


\end{document}